\documentclass{bmvc2k}
\usepackage{graphicx}
\usepackage{amsmath,amssymb} 
\usepackage{color}
\usepackage{tabularx}
\usepackage{mdwlist}
\usepackage[ruled,vlined,noend]{algorithm2e}
\usepackage[percent]{overpic}
\usepackage{ccicons}
\usepackage{hyperref}
\hypersetup{
	urlcolor=,
}
\usepackage{ojw_style}
\definecolor{lightgray}{rgb}{0.85, 0.85, 0.85}

\newcommand{\error}{\mathcal{E}}
\newcommand{\rot}{\mat{R}}
\newcommand{\trans}{\vec{t}}
\newcommand{\sensor}{\vec{s}}
\newcommand{\lens}{\vec{l}}
\newcommand{\nframes}{P}
\newcommand{\ncameras}{C}
\newcommand{\nlandmarks}{L}
\newcommand{\landmark}{\vec{X}}
\newcommand{\plane}{\vec{n}}

\newcommand{\imcoord}{\vec{x}}
\newcommand{\project}{\pi}

\newcommand{\distort}{\varphi}
\newcommand{\undistort}{\varphi^{-1}}
\newcommand{\calib}{\kappa}
\newcommand{\uncalib}{\kappa^{-1}}
\newcommand{\findex}{i}
\newcommand{\update}[1]{\delta\vec{#1}}
\newcommand{\rodrigues}{\Omega}
\newcommand{\patchlen}{N}
\newcommand{\patchcoords}{\mat{P}}
\newcommand{\visibleframes}{\mathcal{V}}
\newcommand{\robustify}{\rho}
\newcommand{\height}{\textrm{H}}
\newcommand{\width}{\textrm{W}}
\newcommand{\patchtransform}{\Psi}
\newcommand{\image}{\im{I}}
\newcommand{\patch}{\bar{\im{I}}}
\newcommand{\Mean}{\mean{\patch}}
\newcommand{\SD}{\sd{\patch}}
\newcommand{\params}{\mathrm{\Theta}}
\newcommand{\sourceindexlist}{I}
\newcommand{\ftofproject}{\Pi^\params}

\newcommand{\Hrcs}{\mat{H}_\mathrm{rcs}}
\newcommand{\grcs}{\vec{g}_\mathrm{rcs}}
\newcommand{\jacc}{\bar{\jac}_k}
\newcommand{\jacs}{\hat{\jac}_k}
\newcommand{\paramsnostruct}{\bar{\params}}
\newcommand{\compose}{\oplus}
\newcommand{\meanpatch}{\boldsymbol{\mu}}
\newcommand{\worldcoords}{\mat{X}}
\newcommand{\reg}{\mathrm{reg}}

\graphicspath{ {./fig/} }
\newlength{\imlen}

\title{Large Scale Photometric Bundle Adjustment} 

\addauthor{Oliver J. Woodford}{o.j.woodford.98@cantab.net}{1}
\addauthor{Edward Rosten}{}{2}

\addinstitution{
	Snap, Inc., Santa Monica.\\
}
\addinstitution{
	Snap Group Ltd., London.\\
}

\runninghead{Woodford, Rosten}{Large Scale Photometric Bundle Adjustment}
\begin{document}
\maketitle

\begin{abstract}
Direct methods have shown promise on visual odometry and SLAM, leading to greater accuracy and robustness over feature-based methods. However, offline \dims{3} reconstruction from internet images has not yet benefited from a joint, photometric optimization over dense geometry and camera parameters. Issues such as the lack of brightness constancy, and the sheer volume of data, make this a more challenging task. This work presents a framework for jointly optimizing millions of scene points and hundreds of camera poses and intrinsics, using a photometric cost that is invariant to local lighting changes. The improvement in metric reconstruction accuracy that it confers over feature-based bundle adjustment is demonstrated on the large-scale Tanks \& Temples benchmark. We further demonstrate qualitative reconstruction improvements on an internet photo collection, with challenging diversity in lighting and camera intrinsics.
\end{abstract}

\begin{figure}[t]
\setlength{\imlen}{0.195\linewidth}
\begin{tabularx}{\linewidth}{@{}*{3}{c@{\extracolsep{\fill}}}c@{}}
	\begin{overpic}[height=\imlen]{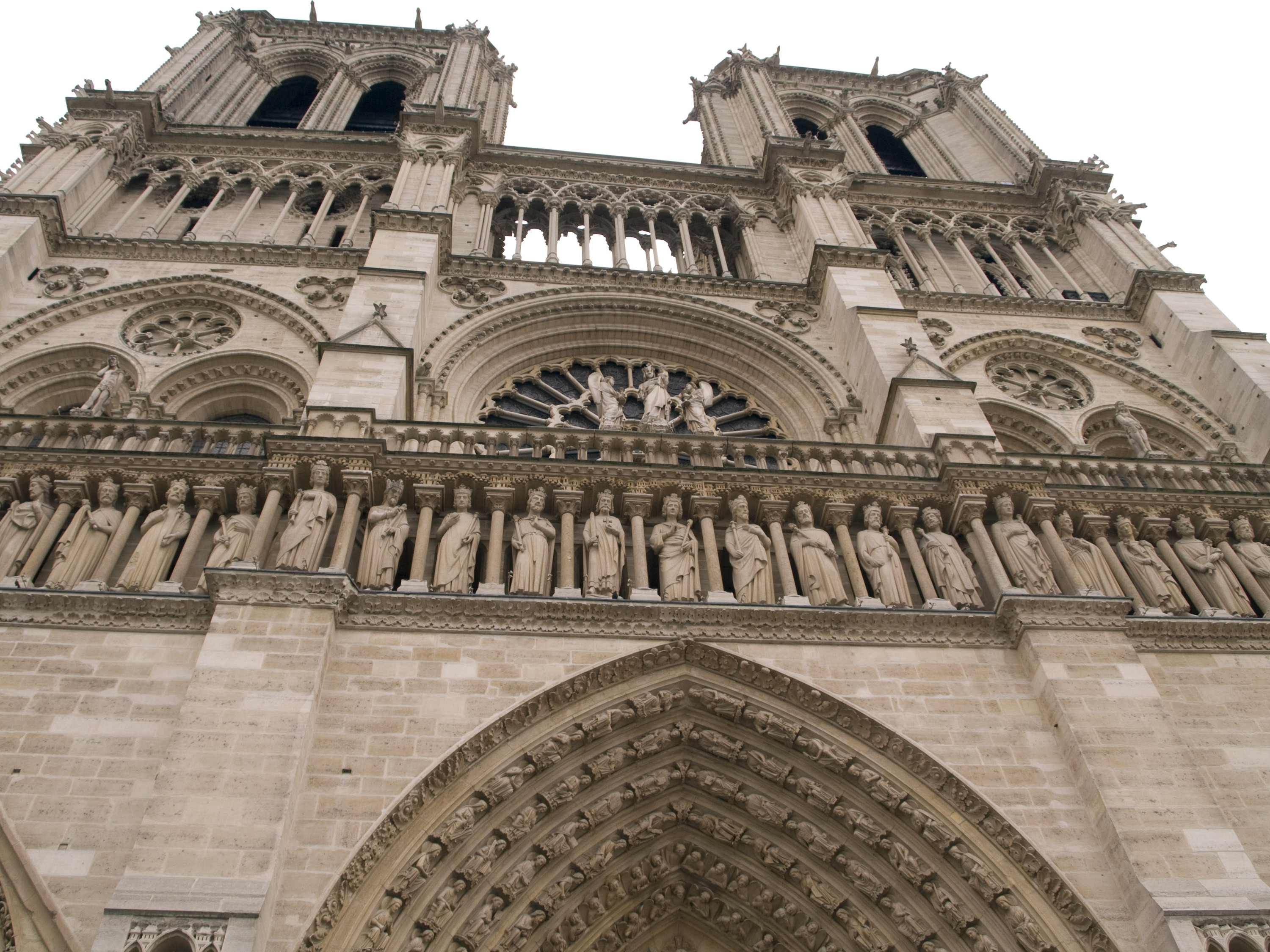}\put(88,66){(a)}\put(1,1){\tiny\color{white} \ccLogo ~\ccAttribution~\href{https://www.flickr.com/photos/hellolapomme/2301160831}{hellolapomme}}\end{overpic} &
	\begin{overpic}[height=\imlen]{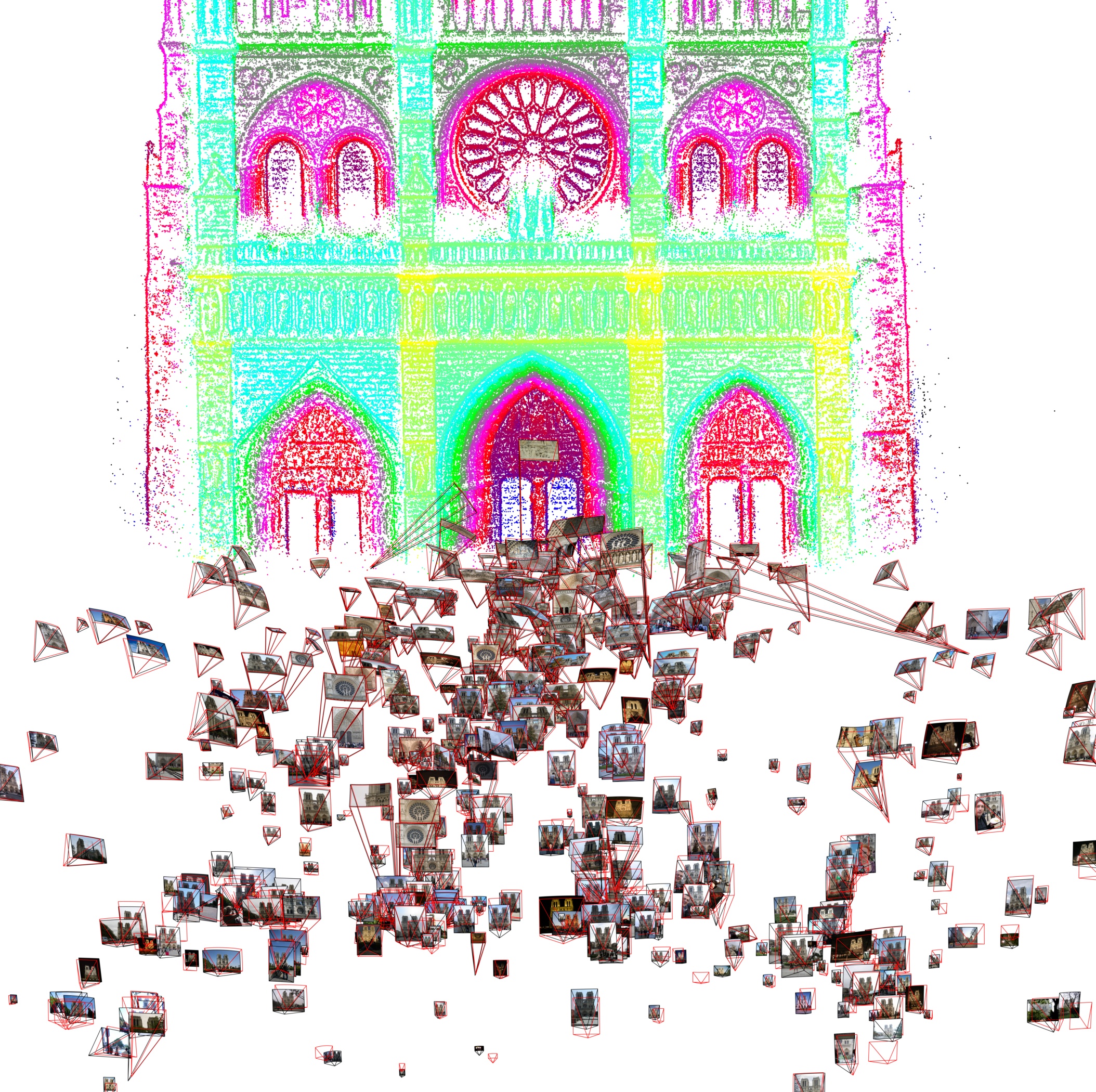}\put(84,89){(b)}\end{overpic} &
	\begin{overpic}[height=\imlen]{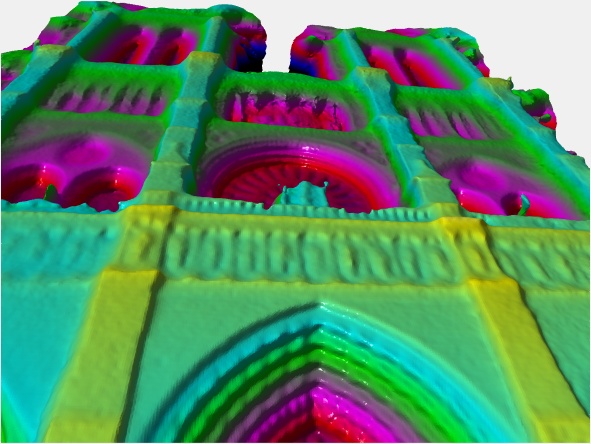}\put(88,66){(c)}\end{overpic}  & 
	\begin{overpic}[height=\imlen]{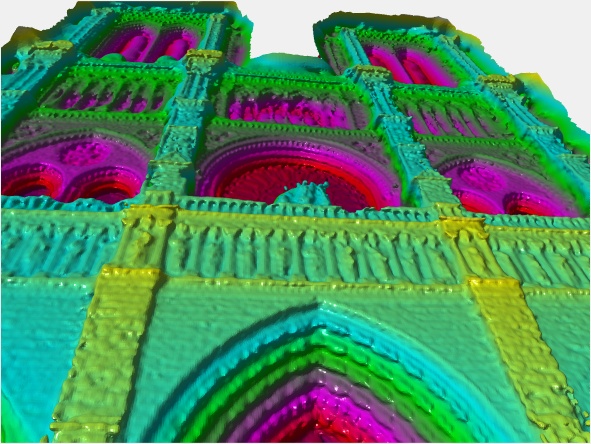}\put(88,66){(d)}\end{overpic}
\end{tabularx}
\caption{\label{fig:notre_dame} Given 700+ photos of Notre Dame (\eg a), captured with different cameras and lighting conditions, 
	our method refines the camera poses (b, \emph{red}), intrinsics, and dense geometry (c) produced by a standard SfM+MVS framework~\cite{schonberger2016colmapsfm,schonberger2016colmapdense}, using a joint, photometric optimization. Both the new poses (b, \emph{black}) and \dims{3} landmarks (b) can be used to generate higher fidelity dense reconstructions of the scene, \eg via Poisson meshing~\cite{kazhdan2013poisson}~(d).
}
\end{figure}

\section{Introduction}
\label{sec:introduction}
The joint estimation of camera parameters and scene structure from a set of images is a fundamental Computer Vision problem, with applications from online camera pose estimation for augmented reality, to large scale reconstruction of objects, buildings and cities for mapping, game asset generation and historical archiving. The former, visual odometry task has recently been shown~\cite{engel2017direct} to significantly improve in accuracy when using a photometric error, rather than the geometric error of more traditional, feature-based methods. There are good theoretical reasons for this: these ``direct'' methods optimize in the domain of pixel errors, the true source of measurement noise. In addition, the approach requires localizability in only \dims{1}, along epipolar lines, rather than \dims{2} for feature-based methods. This enables the use of intensity \emph{edges}, in addition to corners, allowing for a denser reconstruction, and thus more constraints on camera parameters also.

Despite these advantages, the latter task of large scale reconstruction, in particular from sets of internet images, consisting of a large number of photos, each with their own camera intrinsics and lighting conditions, has not yet benefited from a joint, photometric treatment. The de-facto standard approach to this task is to compute camera parameters and sparse geometry using a feature-based structure from motion (SfM) method~\cite{ozyecsil2017survey}, followed by dense geometry reconstruction using a multi-view stereo (MVS) method. The goal of this work is to bring the benefits of a photometric error to the first stage, joint camera and structure estimation, improving the accuracy of inputs to an MVS second stage. Specifically, we tackle the problem of large scale, photometric bundle adjustment, \ie the joint refinement of camera and structure parameters under a photometric error, addressing two key challenges:
\begin{enumerate*}
	\item Handling the variety of both lighting conditions and intrinsic parameters present in a large and diverse set of source images, such as those downloaded from the internet.
	\item Solving an optimization problem involving thousands of camera variables and millions of geometry variables in an efficient and effective manner.
\end{enumerate*}
We do not tackle the initialization problem, required for a fully photometric SfM pipeline, instead using off-the-shelf software to generate initial parameters. 
However, we demonstrate that even a photometric refinement of feature-based estimates yields a significant improvement in reconstruction accuracy, as demonstrated quantitatively using the Tanks and Temples (TT) benchmark~\cite{tanks-temples}, and qualitatively on an internet photo collection.

\section{Related work}
\label{sec:related_work}

The joint optimization of structure and camera parameters is common within feature-based systems~\cite{ozyecsil2017survey}, which minimize a geometric error. Often the feature locations themselves are a result of a photometric optimization (\eg KLT tracking~\cite{lucas1981iterative}).  Alternating optimizations of such geometric and photometric errors yields improvements~\cite{furukawa2009accurate}. However, few methods exist which minimize a photometric error directly over structure and motion. 
These fall into two main categories: offline reconstruction~\cite{delaunoy2014photometric,goldlucke2014super}, and visual odometry (VO)~\cite{alismail2016photometric,alismail2016direct,engel2017direct,kahler2011tracking}.

The offline methods~\cite{delaunoy2014photometric,goldlucke2014super} model scene structure densely with triangulated meshes, regularized for smoothness. A texture map is inferred, using a texture-to-image error, allowing appearance to be super-resolved~\cite{goldlucke2014super}. 
This significantly increases both the number of variables (due to the texture) and dependence between them (due to the mesh and smoothness regularization). As a result, optimization is either alternated over different sets of variables (texture, structure, cameras)~\cite{goldlucke2014super}, or a simple, first order, gradient descent solver~\cite{delaunoy2014photometric}.

The VO methods minimize an image-to-image error~\cite{alismail2016photometric,alismail2016direct,engel2017direct,kahler2011tracking,park2017illumination}, using the structure to compute correspondences between images, thus avoiding the need to infer texture. Image-to-image errors require handling both lens distortion and inverse, or un-, distortion. Camera intrinsics are assumed known and fixed, thus avoiding optimizing lens parameters through the undistortion process. Most methods~\cite{alismail2016photometric,alismail2016direct,engel2017direct,park2017illumination} model structure with sparse, ray-based landmarks: fronto-parallel patches anchored to a pixel location in a source frame, with variable depth. Some MVS methods~\cite{furukawa2010pmvs,habbecke2006plane} optimize both the depth and normal of landmarks, though not jointly with camera parameters. Similarly, earlier photometric VO work~\cite{kahler2011tracking} tracks a few planes of broad extent, optimizing both plane parameters and camera extrinsics. Without smoothness regularization, the landmarks or planes of these VO methods~\cite{alismail2016photometric,alismail2016direct,engel2017direct,kahler2011tracking} are independent of each other. The methods do joint optimization using second order solvers, improving the speed of convergence, but on relatively small problems. 

Most of these methods assume constant brightness of a scene point in all images~\cite{alismail2016photometric,delaunoy2014photometric,engel2017direct,goldlucke2014super,kahler2011tracking}. Non-Lambertian surfaces or lighting changes due to time of day or year, or a shifting light source, or by images taken with different cameras, invalidate this assumption. Alismail \etal~\cite{alismail2016direct} transform images into an 8-channel, lighting invariant, binary feature space prior to minimizing the photometric error; this makes the method invariant to local lighting changes, at a cost to computation time and convergence basin size~\cite{woodford2018ncc}. Park \etal~\cite{park2017illumination} evaluated this and other approaches to illumination robustness in the context of direct SLAM. MVS frameworks often use the Normalized Cross Correlation (NCC) photometric score~\cite{furukawa2010pmvs,schonberger2016colmapdense}, which is invariant to affine intensity variations,  over local patches. Recent work on image alignment~\cite{woodford2018ncc} has incorporated this measure into a standard, least squares optimization framework, employed here.

\subsection{Our contributions}
Despite computing dense geometry, our approach has more in common with the VO approaches mentioned above, using an independent, ray-based, planar landmark representation for structure, and a joint, second order solver for optimization. 
We contribute the following:
\begin{enumerate*}
	\vspace{-1pt}
	\item Applying an NCC-based photometric framework~\cite{woodford2018ncc} to bundle adjustment. While this measure has been applied to both tracking and MVS, it has not been optimized jointly over both structure and camera parameters.
	\item Optimizing lens distortion parameters with image-to-image errors, requiring differentiation through the lens undistortion process.
	\item A memory efficient implementation of the Variable Projection optimizer~\cite{golub1973varpro,hong2016varproba}, enabling the joint optimization of thousands of camera parameters and millions of structure parameters on a desktop PC.
\end{enumerate*}

\section{Method}
\label{sec:method}
We now describe the key components of our framework: parameterization of camera and structure variables, the photometric cost function, and the optimization framework, plus additional implementation details.

\subsection{Problem parameterization}
Camera parameters define the projection of a \dims3 point, $\landmark\in\real^3$, in world coordinates, onto the image plane, in pixel coordinates.
Camera extrinsics consist of $\nframes$ world to image rotations and translations, $\{\rot_i,\trans_i\}_{i=1}^\nframes, \rot_i\in\SO3$, $\trans_i\in\real^3$, one pair per \emph{image}. Intrinsics consist of $\ncameras$ linear and lens calibration parameters, $\{\sensor_j,\lens_j\}_{j=1}^\ncameras$, $\sensor_j\in\real^4$, $\lens_j\in\real^2$, one pair per \emph{camera}, where $\ncameras\le\nframes$. When $\ncameras<\nframes$, some camera intrinsics are shared across input images; in this case an index mapping from image $i$ to camera $j$ is required as input. To simplify notation, we hide this mapping where necessary, and refer to both extrinsics and intrinsics of a given image using the same index.
The world to pixel coordinate ($\imcoord'$) transform is then given by
\begin{equation}
\label{eqn:world2image}
\imcoord' = \calib_{\sensor_j}(\distort_{\lens_j}(\project(\rot_j\landmark + \trans_j))),
\end{equation}
where $\project(\cdot) : \real^3 \rightarrow \real^2$ is the projection function $\pi([x,y,z]^\trsp) = [x/z,y/z]^\trsp$, $\distort_\lens(\cdot) : \real^2 \rightarrow \real^2$ is a lens distortion function, and $\calib_\sensor(\cdot) : \real^2 \rightarrow \real^2$ is the linear calibration function
\begin{align}
\calib_\sensor(\imcoord) = \begin{bmatrix} s_1 & 0 \\ 0 & s_2 \end{bmatrix} \imcoord + \begin{bmatrix} s_3 \\ s_4 \end{bmatrix}, ~~~~~\text{\suchthat}~~~~~
\uncalib_\sensor(\imcoord) &= \begin{bmatrix} 1/s_1 & 0 \\ 0 & 1/s_2 \end{bmatrix} \left(\imcoord - \begin{bmatrix} s_3 \\ s_4 \end{bmatrix}\right).
\end{align}
For lens distortion, we use a standard polynomial radial distortion model:
\begin{equation}
\distort_\lens(\imcoord) = \imcoord(1 + l_1r + l_2r^2 + .. + l_nr^n), ~~~~~\textrm{where}~~~~~r = \|\imcoord\|^2.
\end{equation}
For world to camera distortion, $n=2$ and $[l_1,l_2]=\lens_j$. For camera to world undistortion (required for our ray-based formulation, described below), the same model can be used with a different set of polynomial coefficients representing the inverse transformation, \suchthat $\undistort_\lens(\imcoord)=\distort_{\phi(\lens)}(\imcoord)$. We compute the undistortion coefficients, $\phi(\lens)$, in closed form using the first six coefficient formulae of Drap \& Lef{\`e}vre~\cite[Appendix C]{drap2016lens}.

We use a ray-based parameterization of structure, whereby each landmark is anchored to a pixel in an input image. Since we are comparing image texture around such points, we avoid making assumptions about the normal of the surface, and instead model it explicitly. Each landmark consists of a given (fixed) pixel location $\imcoord$, source frame index $\findex$, and the variable surface plane parameterization $\plane \in \real^3$ of Habbecke \& Kobbelt~\cite{habbecke2006plane}, in the source image coordinate frame. Its world coordinates are then computed as follows:
\begin{equation}
\label{eqn:image2world}
\landmark = \rot_\findex^\trsp\left(\frac{\bar{\imcoord}}{\plane^\trsp\bar{\imcoord}} - \trans_\findex\right), ~~~~~\text{where}~~~~~\bar{\imcoord} = \begin{bmatrix} \undistort_{\lens_i}(\uncalib_{\sensor_i}(\imcoord)) \\ 1 \end{bmatrix}.
\end{equation}
A pixel to pixel correspondence $\imcoord\rightarrow\imcoord'$ from source frame $i$ to target frame $j$, for landmark $k$, can thus be achieved through the substitution of equation (\ref{eqn:image2world}) into equation (\ref{eqn:world2image}), which we represent, for $\patchlen$ image coordinates, constituting a patch around the landmark, with the function $\ftofproject_{ijk}(\cdot) : \real^{2\times\patchlen} \rightarrow \real^{2\times\patchlen}$. $\params=\{\{\rot_i,\trans_i\}_{i=1}^\nframes,\{\sensor_j,\lens_j\}_{j=1}^\ncameras, \{\plane_k\}_{k=1}^\nlandmarks\}$ denotes the set of all problem parameters, the variables to be optimized. $\nlandmarks$ is the number of landmarks.

\subsubsection{Update parameterization}
Each iteration of optimization computes a parameter update, $\delta\params$. Most parameters, with the exception of rotations, minimally parameterize a Euclidean space, therefore are updated additively, \eg $\plane_k \leftarrow \plane_k + \update{\plane}_k$. For rotations, the update is parameterized (minimally) as $\rot_i \leftarrow \rot_i \rodrigues(\update{r}_i)$, where $\rodrigues(\cdot)$ is Rodrigues' formula~\cite{rodriguesformula} for converting a 3-vector into a rotation matrix. In an abuse of notation, by a derivative of rotation, $\frac{\partial}{\partial \rot}$, we mean the derivative of the update, $\frac{\partial}{\partial \update{r}}\big|_{\update{r}=0}$. The update of parameters in general is denoted $\params\leftarrow\params\compose\delta\params$.

\subsection{Cost formulation}
\label{sec:ls_cost}

Our parameterization gives us a mapping from pixels in one image to pixels in another, via the scene geometry and camera positions and intrinsics; our cost should measure the difference between those two sets of pixels. To ensure that our cost is invariant to local lighting changes as well as unexpected occlusions, we use a robust, locally normalized, least squares NCC cost~\cite{woodford2018ncc}. Specifically, for each landmark (indexed by $k$), anchored in image $\image_i \in \real^{\height\times\width}$ (we use grayscale images), where $i = \sourceindexlist_k$ is the source image index of the \kth landmark, we define a $4 \times 4$ patch of pixels centered on it, with the set of image coordinates $\patchcoords_k \in \real^{2\times\patchlen}$ ($\patchlen$ = 16). Each landmark is visible in a subset of input frames, the (given) set of indices of which is denoted $\visibleframes_k$. The cost over all landmarks and images is thus given by
\begin{align}
\label{eqn:cost}
\textit{Total cost:} && 
\E{}(\params) &= \|\error_\reg\|^2 + \sum_k\sum_{j\in\visibleframes_k}\robustify\left(\|\error_{jk}\|^2\right), ~~~~~~~ \robustify(s) = \frac{s}{s+\tau^2},\\
\textit{Patch residual:} && 
\error_{jk} &= \patchtransform\left(\image_j\left(\ftofproject_{ijk}\left(\patchcoords_k\right)\right)\right) - \patchtransform\left(\image_i(\patchcoords_k)\right), ~~~~~~~ i = \sourceindexlist_k, \\
\textit{NCC normalization:} && 
\patchtransform(\patch) &= \frac{\patch - \Mean}{\SD},~~~~~~~~~~~
\Mean = \frac{\onestrsp\,\patch}{\patchlen},~~~~~~~~~~~
\SD = \|\patch - \Mean\|,
\end{align}
with $\image(\patchcoords) = \patch$ representing sampling, $\ones$ denoting a vector of ones, and $\error_\reg$ being a regularization term (see eq.~(\ref{eqn:intrinsic_regularization})) that ensures  camera intrinsics, known to suffer from degeneracies~\cite{brooks1996recovering}, are well constrained. The Geman-McClure kernel~\cite{black1996robust,geman1985bayesian}, $\robustify$, robustifies costs with $\tau = 0.5$. The source frame, $\sourceindexlist_k$, can be ignored in $\visibleframes_k$, since it contributes no error, by construction.

\subsection{Cost optimization}
\label{sec:ls_optimization}
Equation (\ref{eqn:cost}) defines a robustified non-linear least squares cost, for which many solvers exist~\cite{tingleff2004methods}. These generally involve computing the partial derivatives of residual errors, $\error$, \wrt to the optimization variables, known as the Jacobian,\!\!\footnote{Formulae for specific Jacobians of our cost function are not presented. They can be derived straightforwardly, but modern auto-differentiation tools, such as the C++ Jet type~\cite{ceres-solver} (employed here), make implementing these formulae unnecessary.} $\jac=\frac{\partial\error}{\partial\params}$. Standard implementations of such solvers, \eg Ceres Solver~\cite{ceres-solver}, cache the whole Jacobian, which would be close to 3TB for one dataset used here.
It is not surprising that some approaches resort to alternative strategies to optimize this problem~\cite{goldlucke2014super,delaunoy2014photometric}.

However, our problem has a special structure, common to BA: without surface regularization, the landmarks are independent of each other. Enter the Variable Projection (VarPro) method \cite{golub1973varpro,hong2016varproba}, that, using the Schur complement, allows us to construct and solve a small Reduced Camera System (RCS) problem, then solve for the structure using Embedded Point Iterations (EPIs). The RCS involves the set of all problem variables excluding structure variables, which we denote $\paramsnostruct$. The RCS is constructed and solved, using Levenberg-style damping~\cite{levenberg1944method}, as follows~\cite{hong2016varproba}:
\begin{align}
\label{eqn:update_cams}
\delta\paramsnostruct &= -(\Hrcs + \jac_\reg^\trsp \jac_\reg + \lambda\identity)^{-1}(\grcs + \jac_\reg^\trsp \error_\reg),\\
\label{eqn:accum_rcs}
\Hrcs &= \sum_{k=1}^\nlandmarks \jacc^\trsp\left(\identity - \jacs\jacs^+\right)\jacc, & \grcs =& \sum_{k=1}^\nlandmarks \jacc^\trsp\left(\identity - \jacs\jacs^+\right)\error_k,\\
\error_k &= \left[\robustify'(\error_{jk})\error_{jk}\right]_{\forall j\in\visibleframes_k}, & \robustify'(s) =& \frac{\partial}{\partial s}\robustify(s) = \frac{\tau^2}{(s+\tau^2)^2},\\
\jacc &= \left[\robustify'(\error_{jk})\frac{\partial\error_{jk}}{\partial\paramsnostruct}\right]_{\forall j\in\visibleframes_k}, & \jacs =& \left[\robustify'(\error_{jk})\frac{\partial\error_{jk}}{\partial\plane_k}\right]_{\forall j\in\visibleframes_k},\\
\label{eqn:intrinsic_regularization}
\error_\reg &= 10^5\cdot\begin{bmatrix}\frac{s_{1i}-s_{2i}}{s_{1i}+s_{2i}}~\frac{s_{3i}-W_i/2}{\max(W_i, H_i)}~\frac{s_{4i}-H_i/2}{\max(W_i, H_i)} \end{bmatrix}_{\forall i\in\{1,..,\ncameras\}}^\trsp, & \jac_\reg =& \frac{\partial\error_\reg}{\partial\paramsnostruct},
\end{align}
where $\jac^+ = (\jac^\trsp\jac)^{-1}\jac^\trsp$, denoting the matrix pseudo-inverse, $\identity$ is the identity matrix, $\lambda$ is the damping parameter, and $W_i$ \& $H_i$ are the width \& height of the \ith image respectively.
Following the camera parameter update, EPIs are run using a Gauss-Newton update:
\begin{align}
\label{eqn:update_plane}
\update{n}_k &= -\jacs^+\error_k,
\end{align}
until convergence.  Our implementation groups Jacobians per landmark, and sums the reduced system over landmarks (eq. (\ref{eqn:accum_rcs})). While mathematically equivalent to standard VarPro \cite{hong2016varproba,strelow2016varpro}, this explicitly orders the Jacobian computations.  Both the EPIs and construction of the RCS can thus be run over each landmark independently, in parallel. Jacobians for each landmark are not referenced outside these computations, therefore we do not store Jacobians beyond each iteration of the loops over landmarks, slashing the memory requirements of this method.\!\!\footnote{Previous VarPro bundle adjustment methods~\cite{hong2016varproba,strelow2016varpro} do not provide an explicit Jacobian ordering, therefore cannot exploit this memory reduction. Note that this low memory VarPro can be applied to all bundle adjustments, not just our photometric one.} To limit computation time, VarPro is stopped after just ten iterations. The full optimization is described in Algorithm 1.

\subsection{Initialization, and other implementation details}
The framework presented here jointly refines structure and camera parameters of an existing reconstruction, to improve accuracy. Off-the-shelf SfM + MVS systems can provide an initial $\params$. In particular, we use COLMAP~\cite{colmap} with out-of-the-box parameters\footnote{\texttt{colmap automatic\_reconstructor}, with TT datasets using \texttt{--single\_camera}.} to produce initial camera parameters (via SfM~\cite{schonberger2016colmapsfm}) and landmark parameters (via MVS~\cite{schonberger2016colmapdense}).\!\!\footnote{COLMAP outputs the position and normal direction for each landmark, from which our landmark parameterization, $\plane_k$, can be initialized.} Refining \emph{dense} structure jointly with camera parameters in this way might not be needed in some applications, but serves to demonstrate what is feasible with our low memory formulation. Unnecessary background points slow computation, so we manually select landmarks roughly on the object of interest.

In addition to the camera and landmark parameters, our framework needs a source frame index, $\sourceindexlist_k$, and visibilities, $\visibleframes_k$, per landmark; these remain fixed throughout the optimization. We perform a Poisson surface reconstruction~\cite{kazhdan2013poisson,colmap} on the selected landmarks, and use the resulting mesh to compute visibilities: the mesh is rendered into each view as a depth map, landmarks are projected into the view, and their depths compared to the depth map; those that differ by $<1\%$ are deemed visible. In order to avoid selecting a source frame that is a photometric outlier (\eg due to a specularity), $\sourceindexlist_k$ is chosen as the frame whose patch is closest to a robust mean of the visible, normalized patches:
\begin{align}
\label{eqn:select_source}
\sourceindexlist_k &= \argmin_{j\in\visibleframes_k} \| \patch_j - \meanpatch \|^2, ~~~~ \meanpatch = \argmin_{\hat\meanpatch \in \real^\patchlen} \sum_{j\in\visibleframes_k}\robustify\left(\|\patch_j -\hat\meanpatch\|^2\right),\\
\patch_j &=\patchtransform\left(\image_j\left(\calib_{\sensor_j}(\distort_{\lens_j}(\project(\rot_j\worldcoords_k + \trans_j)))\right)\right),
\end{align}
where $\worldcoords_k$ is a $3\times\patchlen$ matrix of world coordinates, a $4\times 4$ grid of points, spaced such that the mean spacing in visible views is 1 pixel, on the plane around the \kth landmark. $\meanpatch$ is computed using iteratively reweighted least squares~\cite{holland1977irls}, starting from the unrobustified mean. To ensure landmarks are only initialized in textured image regions, we remove ones for which $\|\patch_{\sourceindexlist_k}\| < 0.5\patchlen$ (assuming 256 gray levels).

Image pyramids are used to improve convergence. We run the optimization on half size source frames first, followed by full size. Furthermore, to reduce aliasing, target frames are sampled, using bilinear interpolation, at the image pyramid level which produces image samples that are closest to one pixel apart, for each residual $\error_{jk}$. Finally, we optimize structure alone prior to commencing joint optimization at the first resolution.

\begin{figure}[p]
	\resizebox{0.43\linewidth}{!}{
		\begin{minipage}{0.64\linewidth}
			\begin{algorithm}[H]
				\SetAlgoLined
				$\lambda \leftarrow \nlandmarks; ~~ \omega \leftarrow 10;$~~ \# Set damping parameters \\
				Compute initial cost, $S_0 \leftarrow \E{}(\params_0)$, (eq. (\ref{eqn:cost}))\;
				\For{t~=~1:10}{
					$\params_t \leftarrow \params_{t-1};$\\
					\For{k~=~1:\nlandmarks}{
						Add landmark $k$ to RCS (eq. (\ref{eqn:accum_rcs}))
					}
					\nlset{retry} Compute cameras update (eq. (\ref{eqn:update_cams})); \\
					$\paramsnostruct_t \leftarrow \paramsnostruct_t \compose\delta\paramsnostruct$\;
					
					\For{k~=~1:\nlandmarks}{
						\While{cost decreases}{
							Compute landmark $k$ update (eq. (\ref{eqn:update_plane}));\\
							$\plane_{kt} \leftarrow \plane_{kt} + \delta\plane_k$;\\
						}
					}
					Compute new cost, $S_t \leftarrow \E{}(\params_t)$, (eq. (\ref{eqn:cost}))\;
					\eIf{$S_t < S_{t-1}$}{
						$\lambda \leftarrow \lambda/10; ~~ \omega \leftarrow 10;$ \# Reduce damping \\
					}{
						$\params_t \leftarrow \params_{t-1};$\\
						$\lambda \leftarrow \max(\lambda \omega, 10^{-6})$; \# Increase damping \\
						$\omega \leftarrow 2\omega$;\\
						\textrm{go to} $\mathbf{retry}$\;
					}
				}
				\caption{Low memory VarPro optimization}
			\end{algorithm}
	\end{minipage}}
	\begin{minipage}{0.55\textwidth}
		\setlength{\imlen}{143pt}\
		{\footnotesize
			\begin{tabularx}{\textwidth}{@{}*{3}{c@{\extracolsep{\fill}}}c@{}}
				\includegraphics[height=\imlen]{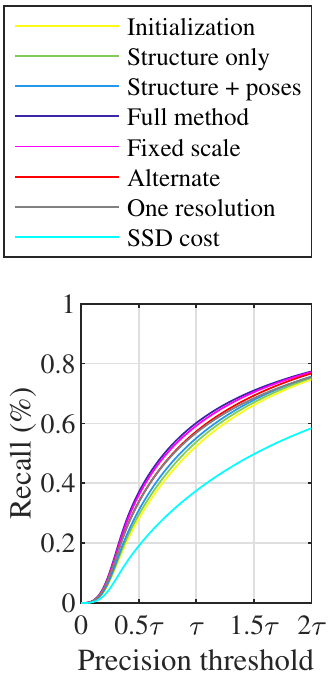} & 
				\includegraphics[height=\imlen]{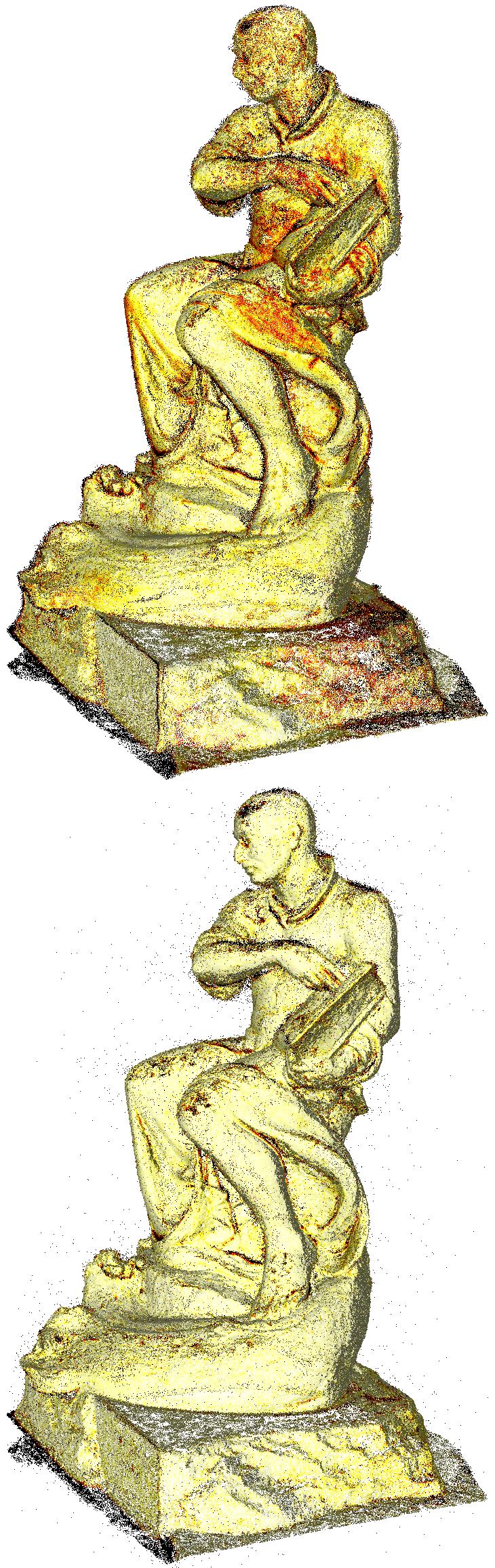} & 
				\includegraphics[height=\imlen]{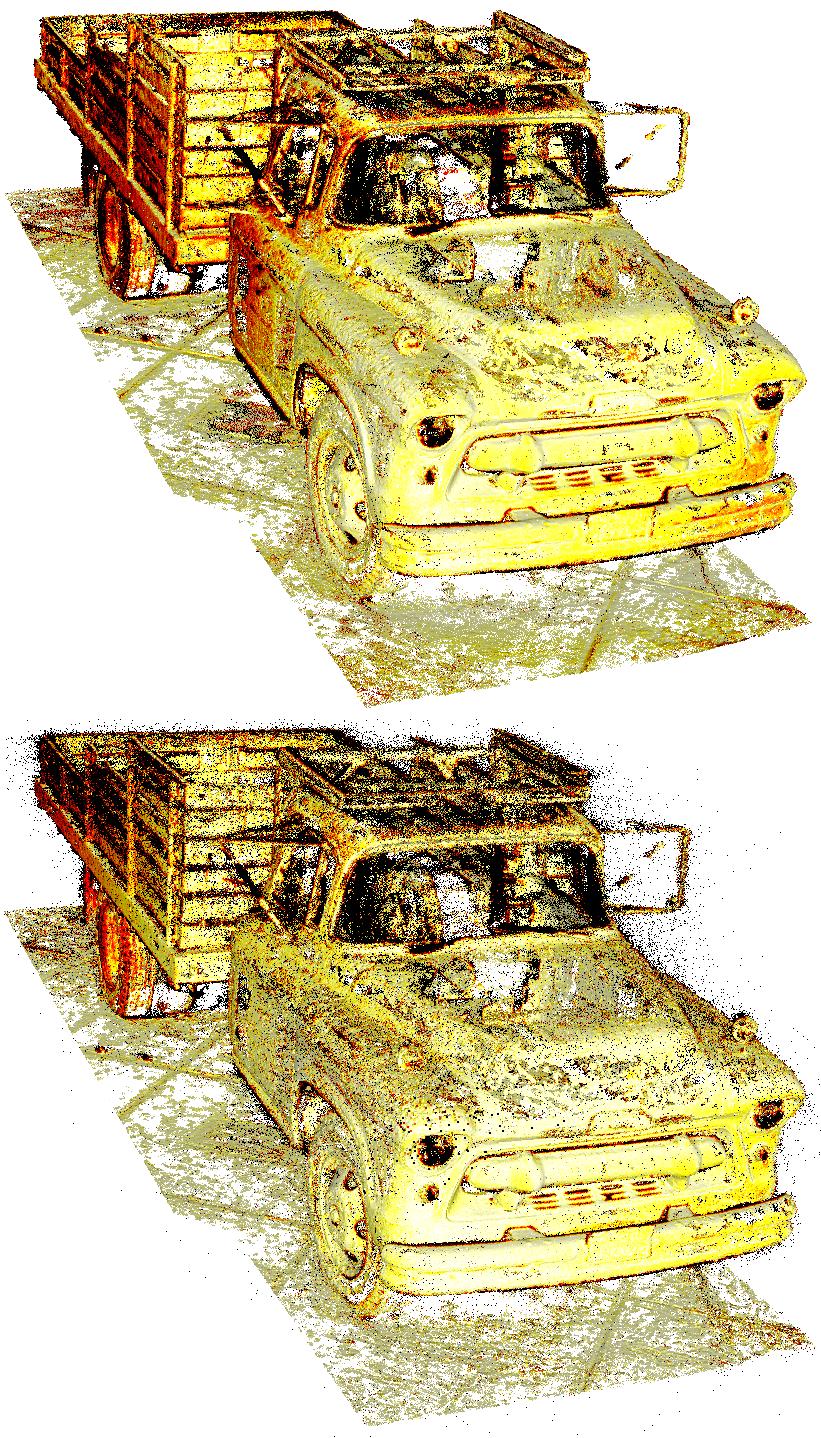} &
				\rotatebox{90}{\hspace{0.19\imlen}LSPBA\hspace{0.3\imlen}COLMAP} \\ 
				(a) Mean precision & (b) Ignatius & (c) Truck
		\end{tabularx}}
		\captionof{figure}{\label{fig:ablation} Results on the TT training sets.}
	\end{minipage}
\end{figure}

\begin{figure}[p]
	\setlength{\imlen}{0.074\linewidth}
	{\tiny
		\begin{tabularx}{\linewidth}{@{}*{9}{c@{\extracolsep{\fill}}}c@{}}
			\rotatebox{90}{~~COLMAP} &
			\includegraphics[height=\imlen]{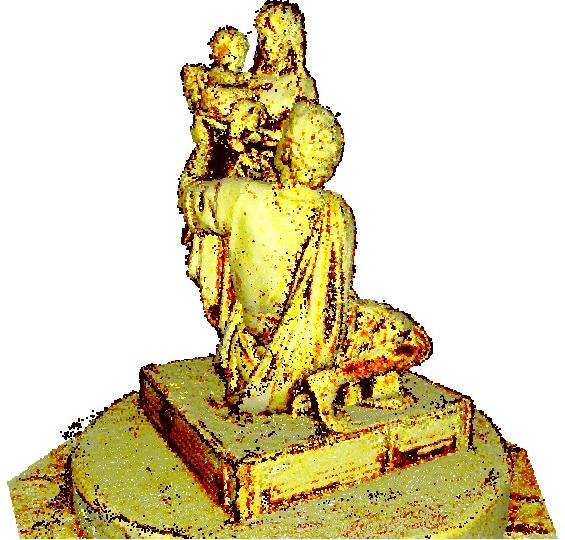} &
			\includegraphics[height=\imlen]{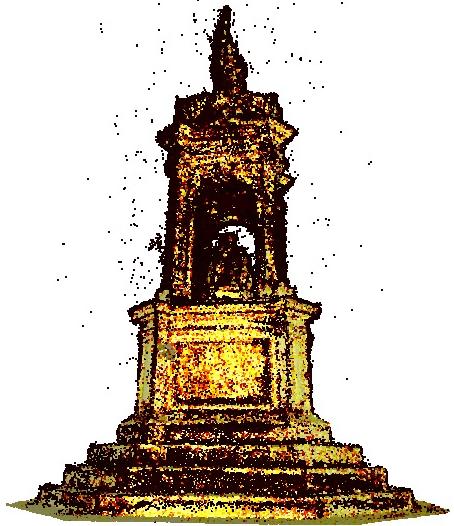} &
			\includegraphics[height=\imlen]{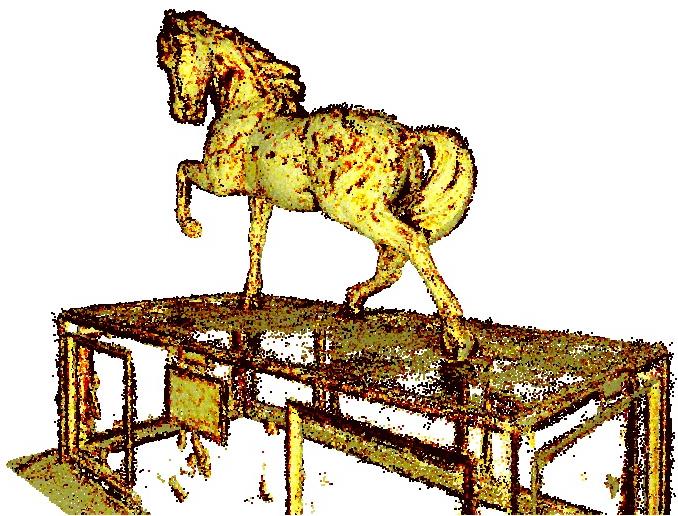} &
			\includegraphics[height=\imlen]{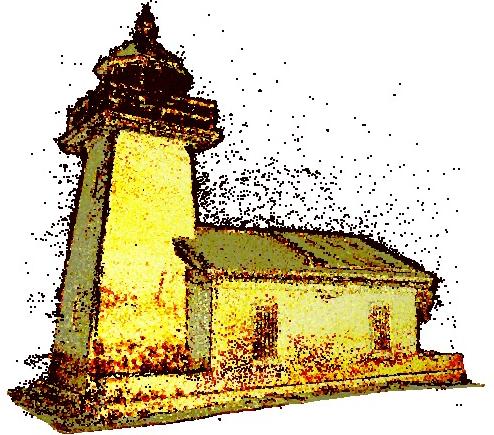} &
			\includegraphics[height=\imlen]{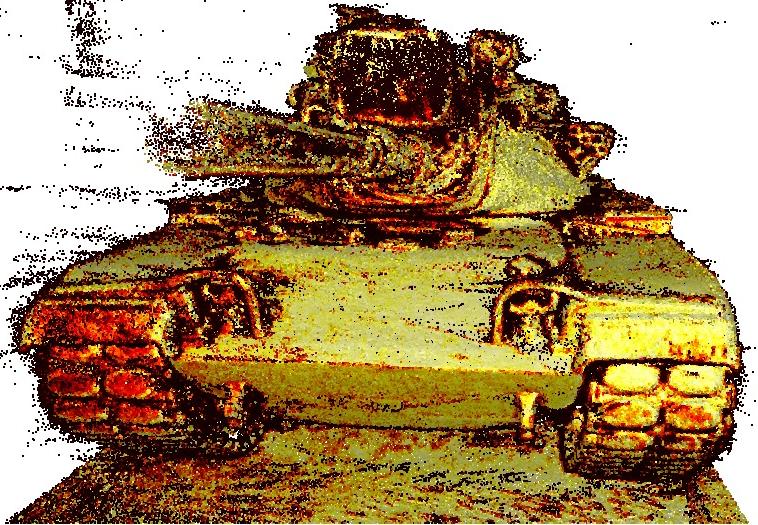} &
			\includegraphics[height=\imlen]{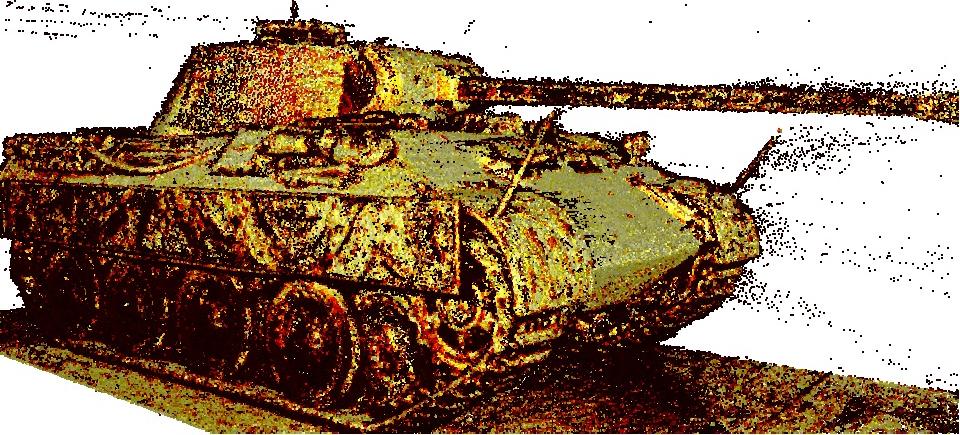} &
			\includegraphics[height=\imlen]{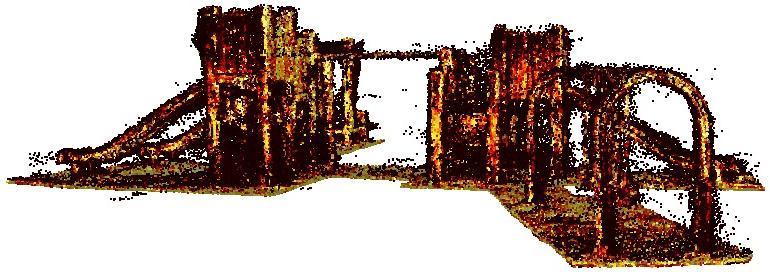} &
			\includegraphics[height=\imlen]{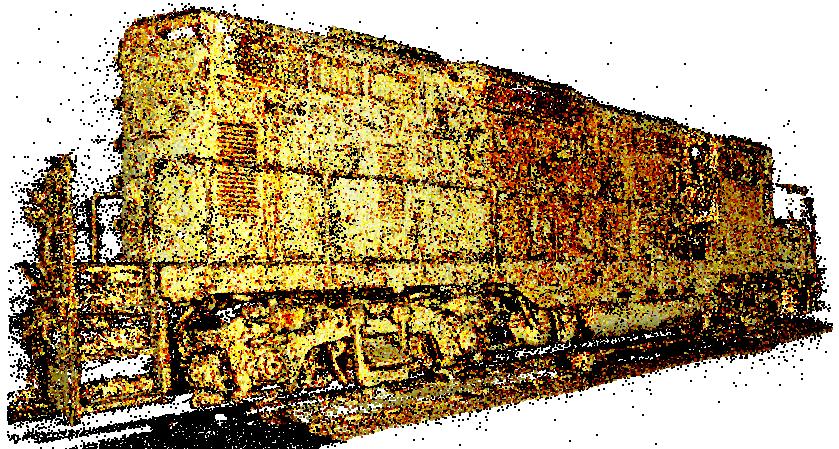} \\
			\rotatebox{90}{~~~~LSPBA} &
			\includegraphics[height=\imlen]{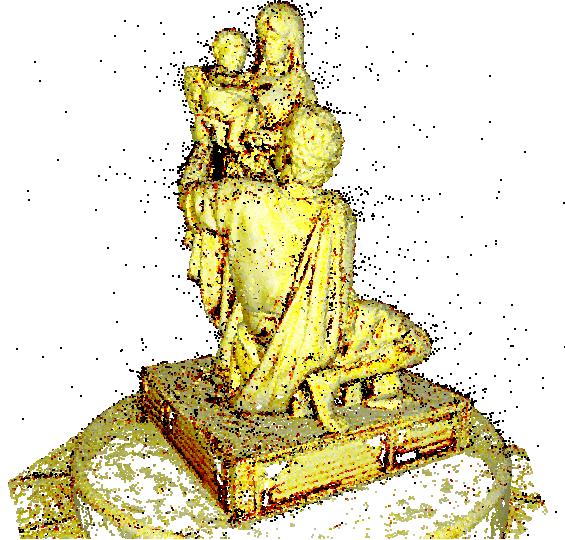} &
			\includegraphics[height=\imlen]{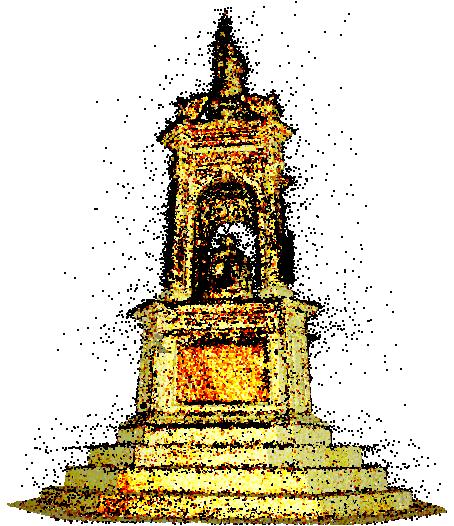} &
			\includegraphics[height=\imlen]{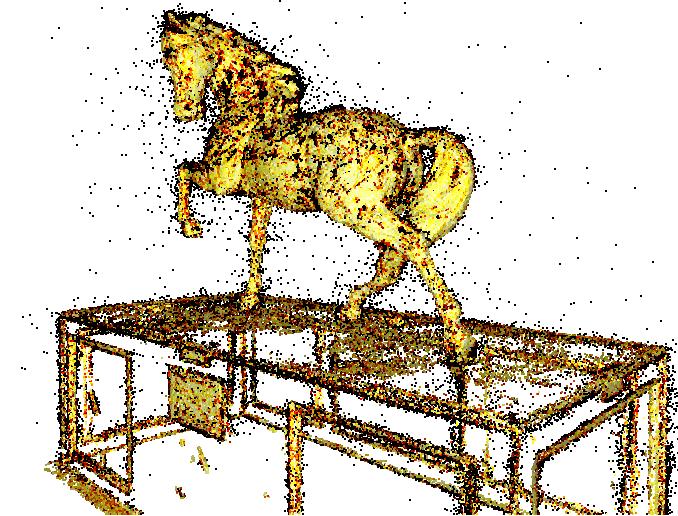} &
			\includegraphics[height=\imlen]{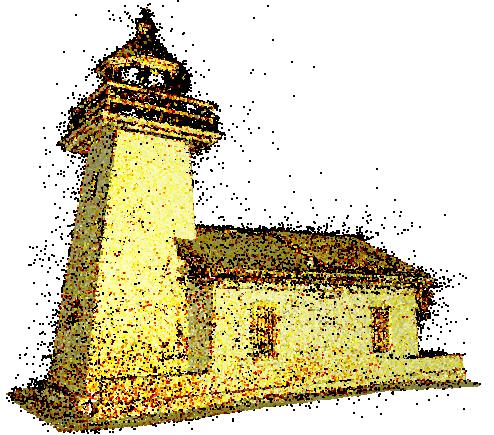} &
			\includegraphics[height=\imlen]{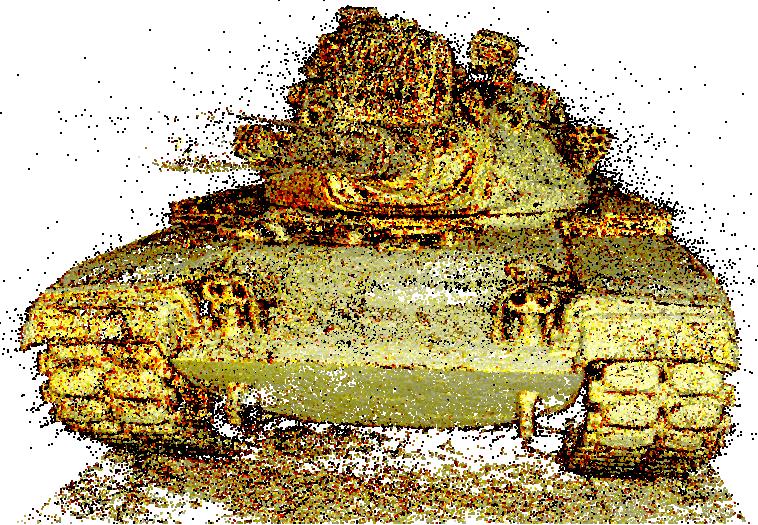} &
			\includegraphics[height=\imlen]{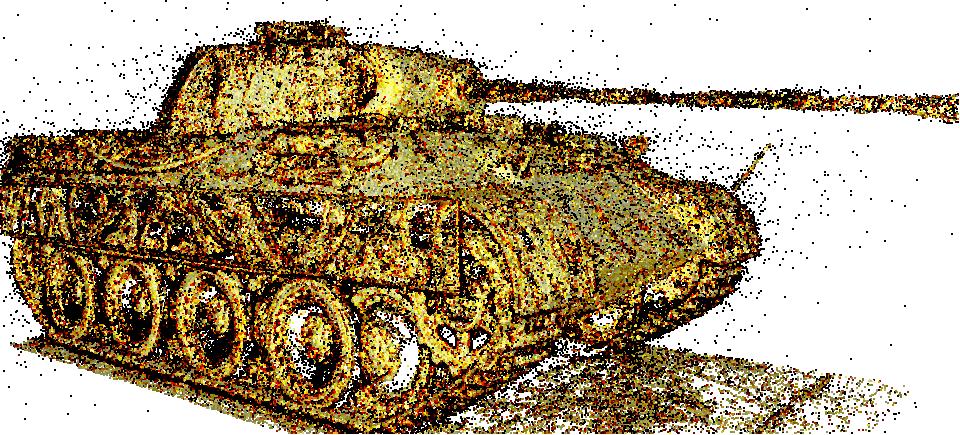} &
			\includegraphics[height=\imlen]{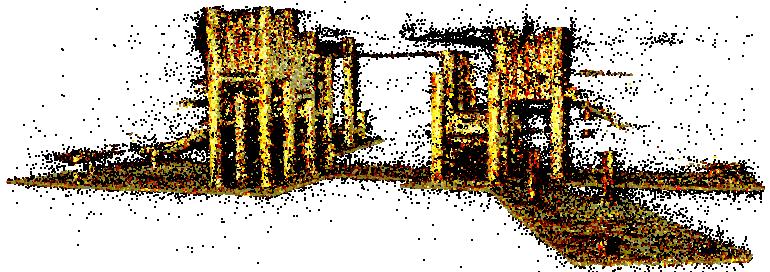} &
			\includegraphics[height=\imlen]{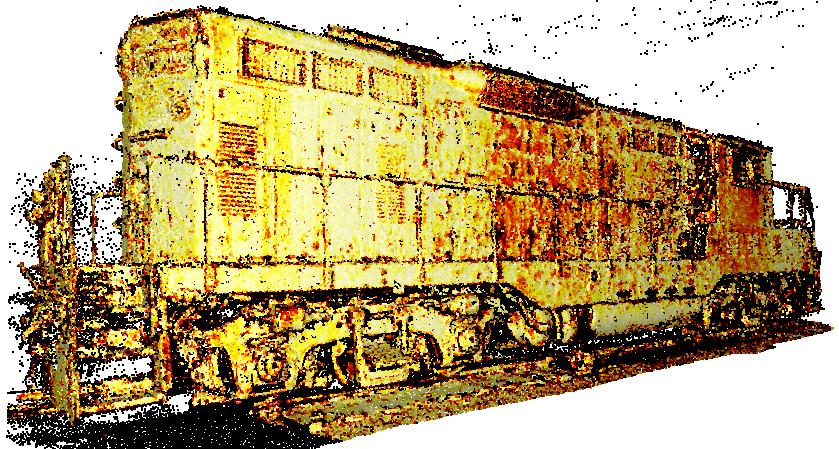} \\
			\scalebox{.7}{\rotatebox{90}{\parbox{1.3\imlen}{\centering LSPBA + COLMAP-MVS}}} &
			\includegraphics[height=\imlen]{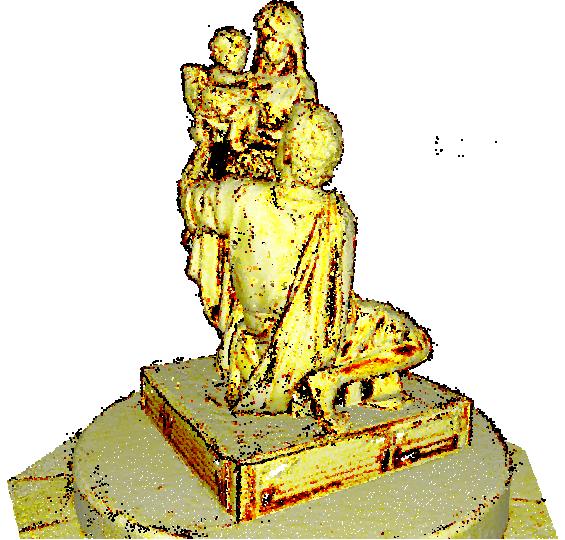} &
			\includegraphics[height=\imlen]{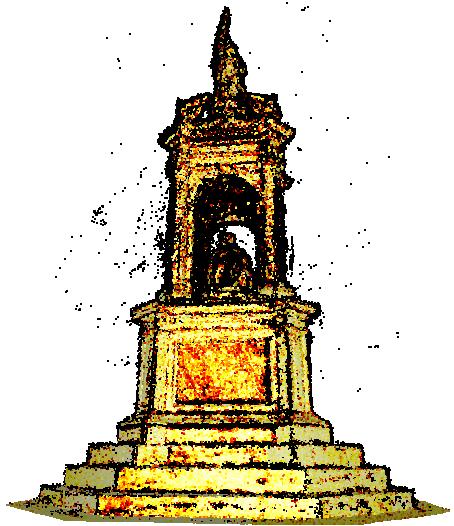} &
			\includegraphics[height=\imlen]{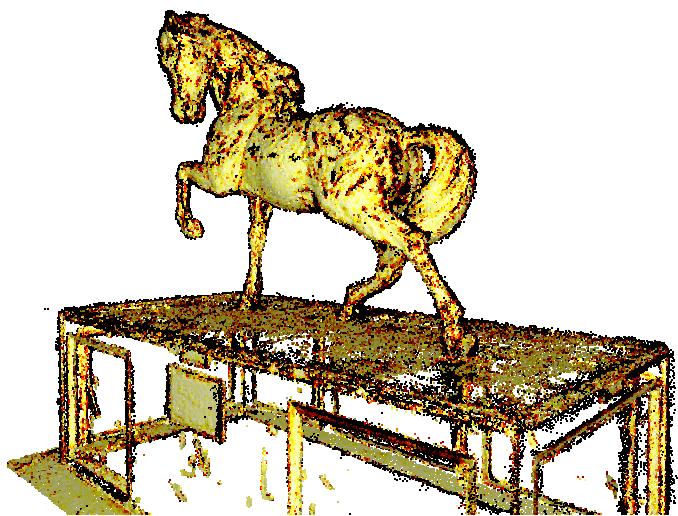} &
			\includegraphics[height=\imlen]{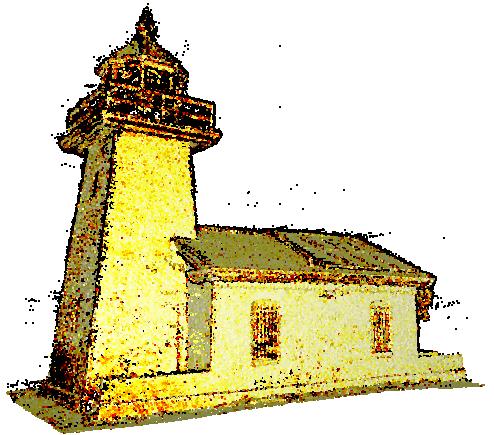} &
			\includegraphics[height=\imlen]{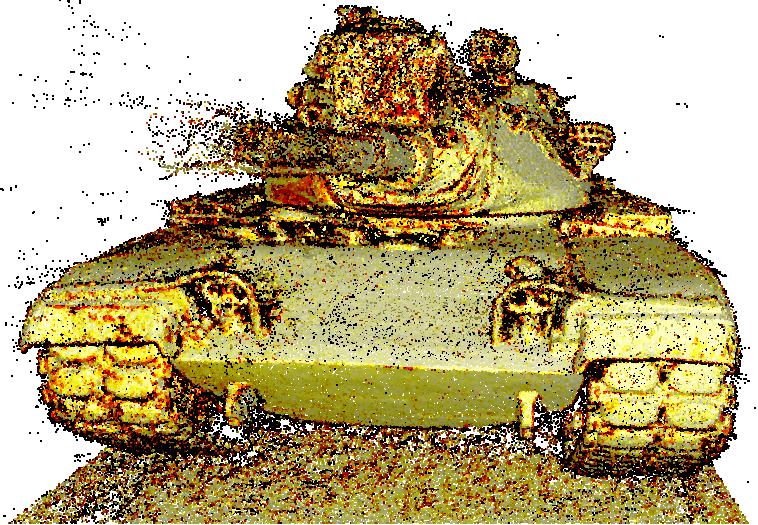} &
			\includegraphics[height=\imlen]{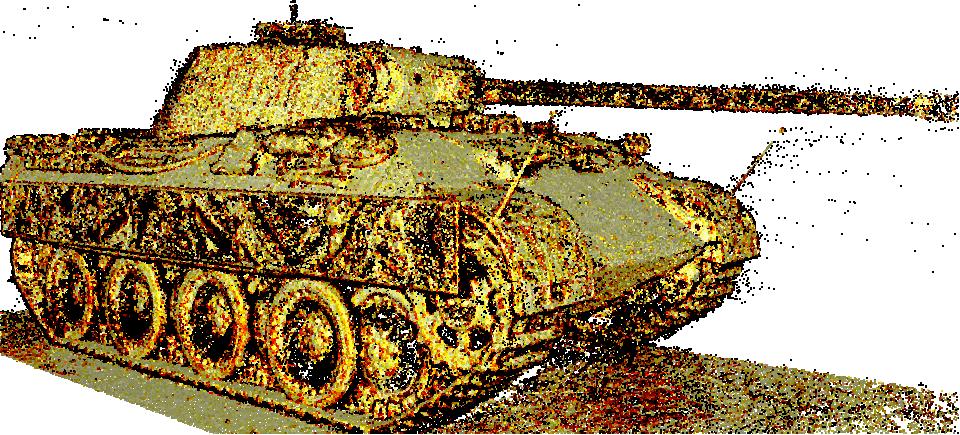} &
			\includegraphics[height=\imlen]{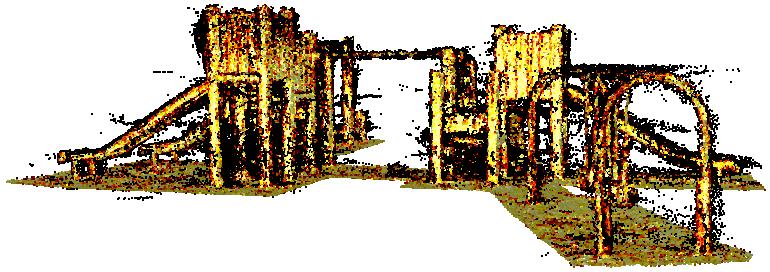} &
			\includegraphics[height=\imlen]{Train_precision_886.jpg} 
	\end{tabularx}}
	\caption{\label{fig:intermediate} Precision error visualization for methods on the TT intermediate sets.}
\end{figure}

\begin{table*}[p]
	\centering
	\resizebox{\linewidth}{!}{
		\begin{tabular}{r||c|c|c|c|c|c|c|c|c}
			& \resizebox{2.3em}{!}{Family} & \resizebox{2.3em}{!}{Francis} & \resizebox{2.3em}{!}{Horse} & \resizebox{2.3em}{!}{\parbox{2.7em}{Light- house}} & M60 & \resizebox{2.3em}{!}{Panther} & \resizebox{2.3em}{!}{\parbox{2.7em}{Play- ground}} & Train & Mean \\
			\hline COLMAP~\cite{schonberger2016colmapsfm,schonberger2016colmapdense} & 56.02 & 34.35 & 40.34 & 53.51 & 41.07 & 39.94 & 38.17 & 41.93 & 43.16 \\
			(Baseline) & \it 45.82 & \it 16.46 & \it 18.79 & \it 59.69 & \it 49.34 & \textbf{\emph{57.01}} & \textbf{\emph{66.61}} & \textbf{\emph{42.15}} & \it 44.48 \\
			\hline LSPBA & \bf 68.76 &  \bf 55.79 & 44.95 & \bf 61.91 & \bf 50.48 & 51.02 & 45.75 & 41.99 & \bf 52.58 \\
			(Our method) & \it 41.48 & \textbf{\emph{26.82}} & \it 15.70 & \textbf{\emph{64.43}} & \it 45.30 & \it 46.45 & \it 53.33 & \it 34.63 & \it 41.48 \\
			\hline LSPBA + & 66.15 & 44.60 &  \bf 45.28 & 57.16 & 50.36 & \bf 51.43 & \bf 48.32 & \bf 43.38 & 50.84 \\
			COLMAP-MVS & \textbf{\emph{55.86}} & \it 23.48 & \textbf{\emph{19.08}} & \it 57.71 & \textbf{\emph{52.50}} & \it 56.15 & \it 64.44 & \it 30.35 & \textbf{\emph{44.95}}
		\end{tabular}
	}
	\caption{\label{tbl:intermediate} Published precision \& \emph{recall} scores for methods on the TT intermediate sets.}
\end{table*}

\begin{table}[p]
	\centering
	\resizebox{\linewidth}{!}{
		\begin{tabular}{r||c|c|c|c|c|c|c|c}
			& Barn & \resizebox{2.4em}{!}{\parbox{2.7em}{Cater- pillar}} & \resizebox{2.4em}{!}{Church} & \resizebox{2.4em}{!}{Ignatius} & \resizebox{2.4em}{!}{\parbox{3.2em}{Meeting- room}} & Truck & Mean & \resizebox{2.4em}{!}{\parbox{2.7em}{Change (\%)}} \\
			\hline COLMAP output & 38.00 & 34.79 & 50.04 & 60.39 & 39.50 & 50.96 & 45.61 & 0 \\
			\hline Structure only & 38.13 & 35.66 & 48.78 & 66.16 & 40.29 & 50.48 & 46.58 & 17.21 \\
			\hline Structure + poses & 41.18 & 35.47 & 47.41 & 69.90 & 40.41 & 50.71 & 47.51 & 33.79 \\
			\hline Full method & \bf 48.00 & \bf 39.39 & 49.42 & 72.61 & \bf 43.00 & \bf 55.00 & \bf 51.24 & 100 \\
			\hline Fixed scale & 45.87 & 39.32 & 48.30 & \bf 72.63 & 42.54 & 54.88 & 50.59 & 88.47 \\
			\hline Alternate & 40.50 & 38.41 & \bf 50.35 & 72.54 & 41.61 & 52.15 & 49.26 & 64.84 \\
			\hline One resolution & 40.50 & 37.76 & 47.51 & 71.98 & 41.08 & 53.66 & 48.75 & 55.74 \\
			\hline SSD cost & 35.77 & 22.28 & 30.43 & 43.86 & 27.42 & 41.75 & 33.58 & -214.0 \\ \hline
			\hline Low qual. initial & 38.35 & 29.76 & 51.91 & 50.63 & 37.26 & 47.55 & 42.58 & - \\
			\hline Low qual. refined & 52.48 & 41.80 & 60.42 & 71.88 & 44.16 & 58.18 & 54.82 & -
		\end{tabular}
	}
	\caption{\label{tbl:ablation}Precision AUC scores for various optimizations (discussed in \S\ref{sec:ablation}) on the TT training sets. The final column shows the percentage increase in score of each method, relative to the increase of LSPBA (full method) over the baseline.}
\end{table}

\section{Evaluation}
\label{sec:evaluation}
While we compute both scene geometry and camera poses, our metric of choice is reconstruction accuracy, rather than the camera position accuracy used by VO methods, since reconstruction is more often the end goal of batch methods. Furthermore, ground truth geometry is more readily available than camera poses on large scale datasets, such as Temples and Tanks (TT)~\cite{tanks-temples}.

We perform a quantitative evaluation of metric reconstruction accuracy (up to scale) using the TT benchmark~\cite{tanks-temples}, whose ground truth geometry was captured by LIDAR. We additionally use their training datasets to run an ablation study highlighting the impact of several elements of our framework. The TT sequences, captured as video from a single camera,\!\!\footnote{For TT sequences we optimize a single, global set of camera intrinsics; for internet photo collections we optimize separate intrinsics for each image.} do not have the variety of lighting conditions and camera intrinsics of an internet-sourced dataset, therefore we also provide qualitative results on an internet photo collection.

Ours is the first \emph{photometric} bundle adjustment method suitable for large, diverse image sets with unknown camera poses and intrinsics; previous approaches have all been feature-based. We therefore pick a baseline from that category: COLMAP (SfM~\cite{schonberger2016colmapsfm} + MVS~\cite{schonberger2016colmapdense})). This method leads publicly available, complete SfM + MVS pipelines on TT in terms of precision (our metric of interest), and it is the initializer for our method, so any difference in performance can be entirely attributed to our framework. Photometric bundle adjustment methods exist for more controlled scenarios~\cite{alismail2016photometric,alismail2016direct,delaunoy2014photometric,engel2017direct,goldlucke2014super,kahler2011tracking}, but the VO methods~\cite{alismail2016photometric,alismail2016direct,engel2017direct,kahler2011tracking} cannot be applied to batches of images, while code is not available for existing batch methods~\cite{delaunoy2014photometric,goldlucke2014super}. Nevertheless, our ablation study contrasts features of our framework with those of other photometric methods, so that our contributions can be fairly evaluated against those. In addition, direct comparisons can be done via the TT online leaderboard. We do not compare to state-of-the-art MVS methods, since they don't optimize camera parameters and also incorporate surface regularization and other prior knowledge.

\subsection{Quantitative precision scores on TT}
We ran our algorithm (LSPBA) on the TT intermediate image sets, and also ran COLMAP-MVS~\cite{schonberger2016colmapdense} using the camera parameters produced by our method (LSPBA + COLMAP-MVS), and submitted both sets of results to the online leaderboard~\cite{tanks-temples}. The resulting scores\footnote{Please refer to the TT paper~\cite{tanks-temples} for details on how the scores are computed.}  are presented in Table \ref{tbl:intermediate}, along with those published for COLMAP.\!\footnote{TT COLMAP results may differ from the initial solutions used here, due to different settings, software versions, stochastic effects, and our culling of landmarks. COLMAP results given in \S\ref{sec:ablation} \emph{are} our initialization (\ie after landmark culling).} Figure  \ref{fig:intermediate} visualizes the reconstructions, with colour encoding the distance from ground truth (lighter is closer).

The LSPBA method significantly improves the metric accuracy of reconstruction over COLMAP, improving the mean precision score by 21.8\%. The recall score is 7\% lower, which is unsurprising given its lack of surface smoothness regularization; this allows some poorly constrained landmarks to leave the surface, particularly visible on the playground sequence. Nevertheless, recall is improved on two sequences. Running an MVS method using the refined camera parameters might be expected to give a similar improvement in accuracy, while maintaining the previous level of recall. This is exactly what LSPBA + COLMAP-MVS achieves; it improves accuracy over COLMAP on every sequence, by 17.8\% on average, whilst slightly improving the average recall also.

\vspace{-3pt}
\subsection{Quantitative ablation study}
\label{sec:ablation}
In order to understand which elements of this framework provide benefit, we ran an ablation study on the TT training image sets, 
resulting in an error-recall curve for precision per sequence, the mean of which is shown in Figure \ref{fig:ablation}(a), where $\tau$ is the sequence dependent error threshold used in the TT benchmark. Also shown in Figure \ref{fig:ablation}(b,c) are error visualizations for the COLMAP (top) and LSPBA (bottom) methods on two sets. The results are summarized in Table \ref{tbl:ablation}, by computing the area under each curve (per sequence), as a percentage of the total plot area. This AUC score captures more information than reporting recall at $\tau$, the value used in the TT benchmark. We describe and discuss each of the results below.

\textbf{Initialization} is the result of COLMAP, with textureless landmarks culled. It is the baseline, and starting point for all the other optimizations. Marginally better than LSPBA (full method) on the Church sequence, it is otherwise significantly worse.

\textbf{Structure only} is a two pyramid level optimization of structure parameters only, keeping camera poses and intrinsics fixed at their initial values. It delivers 17\% of the improvement of the full method, on average, validating the need for a joint optimization.

\textbf{Structure + poses} is a two pyramid level, joint optimization of structure and pose parameters, keeping camera intrinsics fixed at their initial values. It provides 34\% of the total improvement, validating the need to optimize camera intrinsics as well as poses. 

\textbf{Full method} is the complete LSPBA method proposed here; a two pyramid level, joint optimization of structure and camera parameters. It achieves the best score on four of the six sequences, with a significant 12.3\% improvement in AUC over COLMAP.

\textbf{Fixed scale} samples the target image pyramid at the same level as the source image pyramid, rather than using dynamic level selection. Very marginally best on Ignatius, it achieves 89\% of the full method's improvement, demonstrating the modest gains delivered by dynamic level selection.

\textbf{Alternate} replaces the RCS of VarPro with a camera system computed assuming structure is fixed. This then alternates camera and structure updates (10 times), similar to previous work~\cite{furukawa2009accurate,goldlucke2014super}. Marginally best on Church, this approach delivers 65\% of the improvement of the full method overall, validating the benefit of VarPro over alternation.

\textbf{One resolution} applies LSPBA at only the largest image pyramid level, reducing the improvement to 56\% of that using two pyramid levels, demonstrating the benefit of a coarse to fine approach.

\textbf{SSD cost} exchanges the NCC cost of the full method with the sum of squared differences (SSD) cost, which enforces the common constant brightness assumption~\cite{alismail2016photometric,delaunoy2014photometric,engel2017direct,goldlucke2014super}. We used the Huber kernel as robustifier, with a transition threshold of $40^2$\!. This method significantly reduces the precision of the initial solution on all but one sequence, validating the need for a lighting invariant photometric cost in practical applications.

\textbf{Low quality} initial and refined rows refer to using COLMAP on the lowest quality setting\footnote{\texttt{colmap automatic\_reconstructor --single\_camera --quality low}. The density of landmarks is lower (though average precision can be higher), so results are not directly comparable to other rows.} for initialization, and refining this with LSPBA, respectively. Our method improves the accuracy of all sequences, with an average gain in AUC of 29\% (much larger than for the standard initialization), suggesting that it extends well to other initializations.

\subsection{Qualitative results on internet photo collections}
Internet photo collections have a more diverse set of cameras and lighting conditions than the TT datasets, but lack ground truth data. We therefore present only qualitative results, on a publicly available dataset, ``Notre Dame''~\cite{snavely2008modeling}, in Figure \ref{fig:notre_dame}. Panel (b) shows the landmarks, coloured by relief depth, and camera positions before (red) and after (black) refinement. The lowest 10\% of landmarks, ranked by mean photometric cost, are removed to filter out outliers. Comparing the filtered landmarks meshed using Poisson meshing~\cite{kazhdan2013poisson} (d) with the COLMAP landmarks meshed similarly (c), our reconstruction captures significantly finer details, \eg of arches on the towers. It does fail to fix existing, larger scale errors, such as missing balustrade, and introduces more noise on flat regions of the building, due to a lack of texture and smoothness regularization.

To give an idea of the computational resources required for our method, this photo collection, with 701 images and 755k landmarks, took about a week to optimize (not including COLMAP running time), using parallelized\footnote{The two for loops in Algorithm 1 are easily parallelized, \eg using OpenMP.} C++ code on an 8 core 3.7GHz Xeon desktop PC, using 44GB of memory at peak; the full Jacobian for this problem would be 900GB. To accelerate experiments we used a 96 core 3GHz Xeon server; optimization of this dataset took under 5 hours on this machine. This would further improve with GPU acceleration.

\section{Conclusion}
\label{sec:conclusion}
In solving some key challenges, this work enables a new tool for the \dims{3} reconstruction task: refining structure and camera parameters jointly, using a photometric error that is robust to local lighting variations. The framework was evaluated on 15 sets of 150-700 images, with a variety of subject matter. The result is a significant, broad increase in the metric accuracy of reconstruction (up to scale), over a baseline that is representative of the current approach used on this problem: feature-based SfM followed by photometric MVS. Our ablation study provides valuable insight into exactly which aspects of this new approach deliver the most benefit, highlighting the gain in accuracy due specifically to such a refinement.

We have not presented a full system, nor optimized peripheral aspects of the framework, such as landmark selection or visibilities, source frame indices, the robust kernel, landmark weights, or patch sample spacing. We rely on other methods for initialization, which may fail. Improvements are possible in all these areas. Also, we do not propose a replacement to traditional MVS; such systems are complementary, and can be applied after a photometric refinement (which could then use far fewer landmarks), as we show, taking advantage of improved camera pose and intrinsic estimates. We note, however, that our framework could also be incorporated into an MVS method (or surface priors could be added to our method), where all camera parameters are fixed, as well as VO methods, where intrinsic parameters are fixed. Indeed, two widely used MVS frameworks, PMVS~\cite{furukawa2010pmvs} and COLMAP-MVS~\cite{schonberger2016colmapdense}, both use NCC, but neither currently use a second order optimizer or analytic gradients.

\end{document}